\def\paperlanguage{}
\pdfoutput=1 

\documentclass[letterpaper, 10 pt, conference]{ieeeconf}  

\usepackage{bm}
\usepackage{cite}
\include{preamble}

\IEEEoverridecommandlockouts                              

\title{\LARGE \textbf
  {
    \switchlanguage%
    {%
      Continuous Jumping of a Parallel Wire-Driven Monopedal Robot RAMIEL Using Reinforcement Learning
    }%
    {%
      パラレルワイヤ駆動一本脚ロボットRAMIELの\\強化学習による連続跳躍
    }%
  }
}

\author{Kento Kawaharazuka$^{1}$, Temma Suzuki$^{1}$, Kei Okada$^{1}$, and Masayuki Inaba$^{1}$
  \thanks{$^{1}$ The authors are with the Department of Mechano-Informatics, Graduate School of Information Science and Technology, The University of Tokyo, 7-3-1 Hongo, Bunkyo-ku, Tokyo, 113-8656, Japan.
    {\texttt\small [kawaharazuka, t-suzuki, k-okada, inaba]@jsk.t.u-tokyo.ac.jp}
  }
}
\begin{document}

\maketitle
\thispagestyle{empty}
\pagestyle{empty}

\begin{abstract}
  \switchlanguage%
  {%
    We have developed a parallel wire-driven monopedal robot, RAMIEL, which has both speed and power due to the parallel wire mechanism and a long acceleration distance.
    RAMIEL is capable of jumping high and continuously, and so has high performance in traveling.
    On the other hand, one of the drawbacks of a minimal parallel wire-driven robot without joint encoders is that the current joint velocities estimated from the wire lengths oscillate due to the elongation of the wires, making the values unreliable.
    Therefore, despite its high performance, the control of the robot is unstable, and in 10 out of 16 jumps, the robot could only jump up to two times continuously.
    In this study, we propose a method to realize a continuous jumping motion by reinforcement learning in simulation, and its application to the actual robot.
    Because the joint velocities oscillate with the elongation of the wires, they are not used directly, but instead are inferred from the time series of joint angles.
    At the same time, noise that imitates the vibration caused by the elongation of the wires is added for transfer to the actual robot.
    The results show that the system can be applied to the actual robot RAMIEL as well as to the stable continuous jumping motion in simulation.
  }%
  {%
    これまで, パラレルワイヤ機構により速度と力を両立し, 長い加速距離が確保可能な一本脚ロボットRAMIELを開発してきた.
    RAMIELは高い跳躍と連続跳躍が可能であり, 高い走破性能を有している.
    一方で, 関節エンコーダを持たない最小構成のワイヤ駆動型ロボットの欠点として, ワイヤの伸びにより, ワイヤ長から推定した現在の関節速度が振動し, それらの値が信頼できないという問題がある.
    そのため, 高い性能を有しているにも関わらず制御が不安定で, 16回中10回は2回以下の連続跳躍しかできなかった.
    そこで本研究では, シミュレーションにおける強化学習とその実機適用による連続跳躍動作を実現する.
    ワイヤの伸びに伴い振動する関節速度は直接用いず関節角度の時系列から推論を行うと同時に, それらワイヤの伸びによる振動を模したノイズを加えることで実機への転移を行う.
    シミュレーションにおける安定した連続跳躍を実現すると同時に, 実機のRAMIELにおいてもその適用可能性を示した.
  }%
\end{abstract}

\section{INTRODUCTION}\label{sec:introduction}
\begin{figure}[t]
  \centering
  \includegraphics[width=0.7\columnwidth]{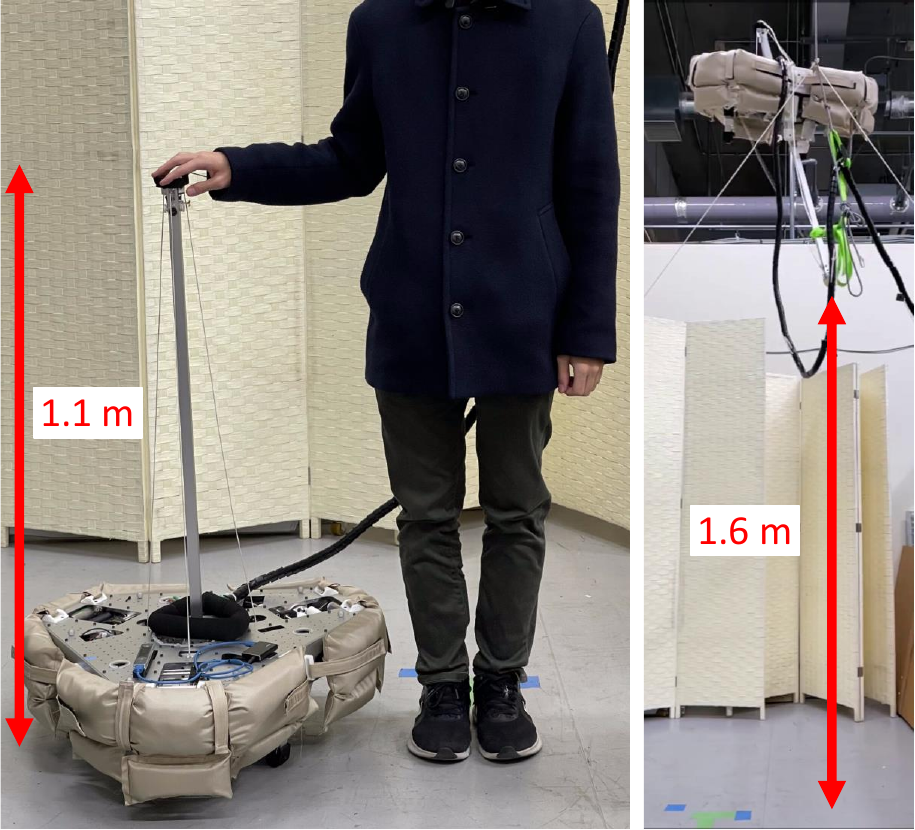}
  \vspace{-1.0ex}
  \caption{Parallel wire-driven monopedal robot RAMIEL \cite{temma2022ramiel}.}
  \vspace{-3.0ex}
  \label{figure:overview}
\end{figure}

\switchlanguage%
{%
  Legged robots have higher running performance than wheeled robots, among which robots that can traverse three-dimensional uneven terrain by jumping have been developed \cite{haldane2017salto1p, kau2019stanforddoggo, kuindersma2016atlas, zaitsev2015taub, matthew2014multimobat}.
  While some robots can make a single jump of about 3 m, continuous jumping is difficult because the actuators required for posture control are reduced \cite{zaitsev2015taub, matthew2014multimobat} (note that in this study, the jump height is defined as the maximum difference between the center of gravity height at takeoff and when jumping).
  On the other hand, robots such as \cite{haldane2017salto1p, kau2019stanforddoggo, kuindersma2016atlas} can jump continuously, but these robots have a maximum jump height of about 1.1 m \cite{haldane2017salto1p}, which is inferior to robots with a single jump.
  Therefore, we have developed a parallel wire-driven monopedal robot RAMIEL (paRAllel wire-driven Monopedal agIlE Leg) that is capable of high and continuous jumps \cite{temma2022ramiel} (\figref{figure:overview}).
  By taking advantage of the linear motion of the actuator in the wire-driven robot, RAMIEL has high-speed, high-power linear motion capability and long acceleration distance, ensuring its jumping performance.
  In previous experiments, RAMIEL has succeeded in a high jump of 1.6 m and a maximum of eight consecutive jumps.

  However, its performance is not yet ideal, especially for the continuous jump.
  This is due to a problem unique to the minimal parallel wire-driven robot.
  In order to keep the joint structure simple and lightweight, the robot does not have a joint encoder, and the joint angle is estimated from the wire length through Extended Kalman Filter \cite{ookubo2015learning}.
  However, since the wire has some elasticity and stretches, the wire length often oscillates in tasks that require large force such as jumping, and as a result, the estimated joint angle often oscillates.
  Therefore, when a general controller \cite{raibert1984onelegged} is applied to RAMIEL, the jumping motion is unstable, and 10 out of 16 trials resulted in less than two consecutive jumps \cite{temma2022ramiel}.
}%
{%
  脚を持つロボットは台車型ロボットに比べ高い走破性能を持つが, その中でもジャンプにより3次元的な不整地を踏破することが可能なロボットが開発されてきている\cite{haldane2017salto1p, kau2019stanforddoggo, kuindersma2016atlas, zaitsev2015taub, matthew2014multimobat}.
  一部のロボットは約3m程度の単発ジャンプが可能な一方, 姿勢制御に必要なアクチュエータを削いでおり連続跳躍が難しい\cite{zaitsev2015taub, matthew2014multimobat} (なお, 本研究ではジャンプ高さを離陸時と跳躍時の重心高さの最大差分と定義している).
  これに対して, \cite{haldane2017salto1p, kau2019stanforddoggo, kuindersma2016atlas}等の連続跳躍が可能なロボットが開発されているが, これらは最大で1.1 m程度\cite{haldane2017salto1p}のジャンプと, 単一ジャンプを志向したロボットよりも跳躍力で劣る.
  そこで, 我々は高跳躍と連続跳躍が可能なパラレルワイヤ型一本脚ロボットRAMIEL (paRAllel wire-driven Monopedal agIlE Leg)を開発してきた\cite{temma2022ramiel} (\figref{figure:overview}).
  ワイヤ駆動型におけるアクチュエータの直動運動を活かすことで, 高速高出力な直線運動能力と長い加速距離を有し, その跳躍性能を確保している.
  実験では, これまで1.6 mの高跳躍と最大で8回の連続跳躍に成功している.

  一方で, 特に連続跳躍において, その性能は高いとは言い難い.
  これは, 最小構成のパラレルワイヤ駆動型ロボット特有の問題に起因する.
  関節構造を簡易で軽量にするために, 関節エンコーダを持たず, ワイヤ長からExtende Kalman Filter等を通して関節角度を推定している\cite{ookubo2015learning}.
  しかし, ワイヤには多少の弾性があり伸びるため, 跳躍のように大きな力が必要なタスクにおいてはワイヤ長が振動, その結果として関節角度推定値が振動する場合が多い.
  そのため, 一般的な\cite{raibert1984onelegged}の制御器をRAMIELに適用した場合, その跳躍動作は非常に不安定であり, 16回中10回は2回以下の連続跳躍しかできなかった.
}%

\begin{figure*}[t]
  \centering
  \includegraphics[width=1.9\columnwidth]{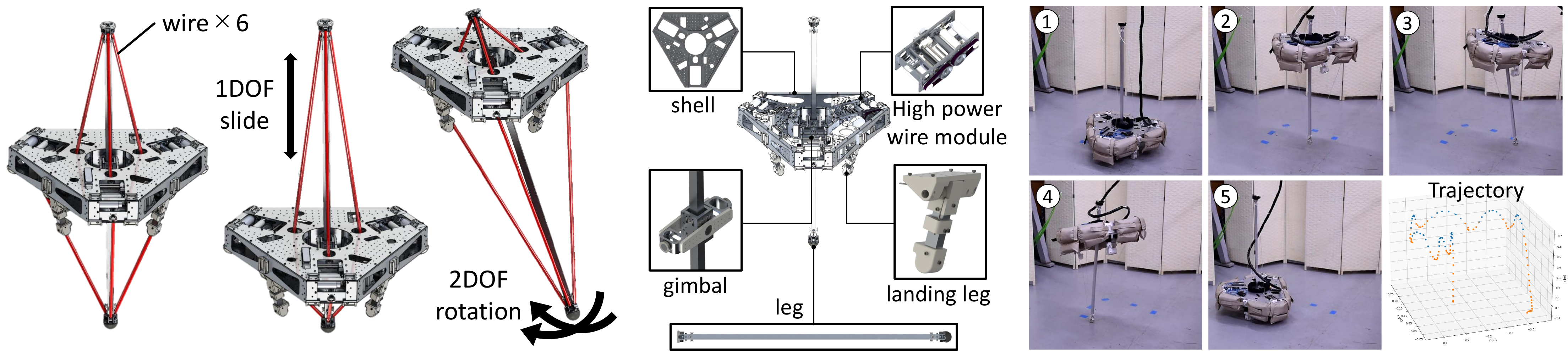}
  \vspace{-1.0ex}
  \caption{Detailed body structure and performance of continuous jumping of RAMIEL.}
  \vspace{-3.0ex}
  \label{figure:ramiel}
\end{figure*}

\switchlanguage%
{%
  Therefore, the objective of this study is to develop a stable continuous jumping motion by reinforcement learning in simulation and transfer it to the actual robot.
  Various reinforcement learning methods have been developed for learning dynamic behaviors of robots \cite{schulman2017ppo, yang2019legged, miki2022anymal, diamond2014reaching, marjaninejad2019tendon}.
  Some of them have been applied not only to simulations but also to actual robots \cite{yang2019legged, miki2022anymal}.
  Reinforcement learning has also been introduced to some wire-driven robots \cite{diamond2014reaching, marjaninejad2019tendon, grimshaw2021cable}.
  On the other hand, \cite{diamond2014reaching, grimshaw2021cable} is applied only to simulations, and \cite{marjaninejad2019tendon} deals only with a simple and relatively static motion of walking with a fixed upper end.
  In this study, we use reinforcement learning to realize dynamic and difficult jumping motions in a wire-driven robot in simulation and on the actual robot, which are more difficult than those in previous studies on manipulators, quadruped robots, etc.
  We show how this can be done by sharing know-how such as the design of the reward function for jumping, the state design without using the velocity term (which is vulnerable to noise due to vibration), the method of adding noise, and the system configuration for application to the actual robot.
  Experimental results show that the simulation results are more stable than those of an existing method, and that our method is applicable to actual robots.
  We believe that this study is an example of the realization of dynamic motion using reinforcement learning in a wire-driven robot with wire stretching, and that future robot configurations with both speed and power will be developed by taking advantage of the wire-driven robot.
}%
{%
  そこで本研究では, シミュレーションにおける強化学習と実機への転移による安定した連続跳躍動作を目的とする.
  これまで様々な強化学習によるロボットの動的動作学習手法が開発されてきた\cite{schulman2017ppo, yang2019legged, miki2022anymal, diamond2014reaching, marjaninejad2019tendon}.
  その中でにはシミュレーションのみではなく, 実機にまで適用した例も存在している\cite{yang2019legged, miki2022anymal}.
  ワイヤ駆動型のロボットについても, 一部強化学習を導入した例もある\cite{diamond2014reaching, marjaninejad2019tendon, grimshaw2021cable}.
  一方で, \cite{diamond2014reaching, grimshaw2021cable}はシミュレーションのみへの適用, \cite{marjaninejad2019tendon}は上端を固定した歩行という単純で比較的静的な動作のみを扱っている.
  本研究では, 強化学習を用いて, ワイヤ駆動型ロボットにおいて, マニピュレータや4脚ロボット等のこれまでの研究にはないほど, 動的で困難な跳躍動作をシミュレーションと実機で実現する.
  その際, 跳躍に向けた報酬関数設計, 振動によるノイズに弱い速度項を用いない状態設計, 実機適用に向けたノイズの加え方とシステム構成等, そのノウハウを示す.
  実験では, シミュレーションにおいて既存手法よりも安定した跳躍が可能であること, また, その実機適用性を示すことに成功している.
  本研究がワイヤの伸びを伴うワイヤ駆動ロボットにおける強化学習を用いた動的動作実現の一例となり, 今後ワイヤ駆動の利点を活かしスピードと力を両立したロボット構成が発展することを確信する.
}%

\section{Parallel Wire-Driven Monopedal Robot RAMIEL} \label{sec:ramiel}
\switchlanguage%
{%
  We describe the body configuration and performance of the parallel wire-driven monopedal robot RAMIEL \cite{temma2022ramiel} handled in this study.
  An overview is shown in \figref{figure:ramiel}.
}%
{%
  本研究で扱うパラレルワイヤ一本脚ロボットRAMIEL \cite{temma2022ramiel}の身体構成とそのパフォーマンスについて述べる.
  その概要を\figref{figure:ramiel}に示す.
}%

\subsection{Structure of RAMIEL}
\switchlanguage%
{%
  \subsubsection{Body Configuration of RAMIEL}
  RAMIEL has an overall height of 1.07 m, an overall width of 0.55 m, and a weight of 10.3 kg.
  The body is mainly divided into the main body and the leg.
  The leg is made only of aluminum square pipes, fixed wire parts, and rubber parts at the ground contact points, and is very light at 0.5 kg.
  This makes the leg highly controllable and suppresses impact when landing.
  The main body has a high-power wire module to control the wires (to be explained later), and landing legs with built-in springs to soften the impact at the time of landing.
  Between the main body and the landing legs, the robot has a gimbal mechanism with two degrees of freedom (DOFs) of rotation (roll and pitch) and one DOF of linear motion, for a total of three DOFs.
  The main body is surrounded by a shock-absorbing exterior with an air damper.

  \subsubsection{High-Power Wire Module}
  RAMIEL is an antagonistic wire-driven robot that drives 3-DOF joints by winding 6 wires.
  The motor is a Maxon BLDC EC-4pole 200W motor, which is decelerated by a belt at a reduction ratio of 2.5:1 to provide sufficient force and speed.
  The motor is driven by a motor driver that can operate at an input voltage of 70V and a maximum instantaneous current of 50A \cite{sugai2018driver}.
  Therefore, a high speed and high torque with a maximum tension of 230 N and a maximum wire winding speed of 10.7 m/sec can be realized, and at the same time, the low gear ratio ensures backdrivability and can cope with impact at landing.
  This output ideally corresponds to a jump height of about 5.8 m.
  In addition, the wires are aligned and wound on the pulleys by the level winder mechanism and the wire holding mechanism, which prevents the wires from overlapping on the pulleys and ensures accurate measurement of wire length and wire tension.
}%
{%
  \subsubsection{Body Configuration of RAMIEL}
  RAMIELの全高は1.07 m, 全幅は0.55 m, 重量は10.3 kgである.
  リンクは主に本体と脚部に分かれている.
  特に脚部はアルミ角パイプとワイヤ固定パーツ, 接地点のゴムパーツのみでできており, 0.5 kgと非常に軽い.
  そのため制御性が高く着地時の衝撃を抑えられる.
  本体にはワイヤを制御する大出力ワイヤモジュール(後に説明する), 着地時の衝撃を和らげるバネを内蔵した着地脚を持つ.
  本体と脚部の間にはロールとピッチの回転2自由度を持つジンバル機構, 直動1自由度の計3自由度を持つ.
  また, 本体の周りには空気ダンパによる衝撃吸収外装が備え付けられている.

  \subsubsection{High-Power Wire Module}
  RAMIELは6本のワイヤを巻き取ることで3自由度の関節を駆動する拮抗ワイヤ駆動型のロボットである.
  十分な力と速度が出せるよう, モータはMaxonのBLDC EC-4pole 200Wを用いており, これがベルトにより2.5:1の減速比で減速されている.
  このモータを, 入力電圧70V最大瞬間電流50Aという条件で動作可能なモータドライバ\cite{sugai2018driver}により駆動する.
  そのため, 最大張力230 N, 最大ワイヤ巻取り速度10.7 m/secという高速高トルクを実現可能であると同時に, 低いギア比によってバックドライバビリティを確保し, 着地時の衝撃に対処することができる.
  なお, 本出力は理想的には重心跳躍高さ約5.8 mに相当する.
  また, ワイヤはレベルワインダ機構とワイヤ押さえ機構によりプーリに整列して巻き取られるため, ワイヤがプーリ上で重なることを防ぎ, 正確なワイヤ長・ワイヤ張力の測定を担保する.
}%

\subsection{Performance of RAMIEL}
\switchlanguage%
{%
  In \cite{temma2022ramiel}, the motion control of the jumping is performed by a controller similar to that used in \cite{raibert1984onelegged}.
  The motion during jumping is divided into two phases: the stance phase in which the landing legs are on the ground, and the flight phase.
  In the stance phase, the two rotational DOFs are controlled so that the body posture becomes horizontal, and the linear DOF is controlled by the energy shaping control law so that the target energy is satisfied.
  In the flight phase, the linear DOF is kept at a certain position, and the two rotational DOFs are controlled by determining the landing point of the target leg based on the horizontal velocity of the robot.

  In addition to the 1.6 m jump shown in \figref{figure:overview}, the robot has successfully made up to 8 consecutive jumps as shown in the right figure of \figref{figure:ramiel} with this control law.
  On the other hand, 10 out of 16 trials resulted in less than two consecutive jumps, which is not a high success rate.
  The wires used to drive the robot in this study are made of a chemical fiber called Zylon{\textsuperscript\textregistered}.
  Zylon{\textsuperscript\textregistered} has high tensile strength, but it also has some elasticity and can stretch by up to 2.5\%.
  In addition, the wire can transmit force only in the tensile direction, and when force is applied in the opposite direction, the tension instantly decreases to zero and the wire loosens, which causes vibration.
  Therefore, sensor values and state estimation are prone to oscillation, which is considered to be the cause of instability in control.
}%
{%
  \cite{temma2022ramiel}では, Raibertらと同様の制御器\cite{raibert1984onelegged}によって動作制御を行った.
  跳躍時の動作を脚が地面に接地しているstance phaseとflight phaseに分けて考える.
  stance phaseでは, 体の姿勢が水平になるように回転2自由度を制御し, 設定した目標エネルギーを満たすように直動自由度をエナジーシェーピング制御則により制御する.
  flight phaseでは, 直動自由度についてはある一定の位置を保ちつつ, 回転2自由度についてはロボットの水平速度から目標の脚の着地地点を決定しこれを実現する.

  \figref{figure:overview}に示す1.6 mのジャンプはもちろんのこと, 本制御則により\figref{figure:ramiel}の右図に示すような最大8回の連続跳躍に成功している.
  一方で, 16回の試行中10回は2回以下のジャンプに留まっており, その成功率は高いとは言い難い.
  本研究でロボットの駆動に用いるワイヤはZylon{\textsuperscript\textregistered}という化学繊維を使用している.
  Zylon{\textsuperscript\textregistered}は高い引張強度を有する一方で, 多少の弾性を有し, 最大で2.5\%程度伸びる.
  また, ワイヤの特性上引張方向にしか力を伝達できず, 逆方向に力が加わると一瞬で張力が0になり緩むという性質を持つため, 振動が起こりやすい.
  そのため, センサ値や状態推定値が振動しやすく, これが制御における不安定さの原因であると考えられる.
}%

\section{Continuous Jumping Motion of RAMIEL Using Reinforcement Learning} \label{sec:proposed}
\switchlanguage%
{%
  The overall system configuration and the definitions of action, state, and reward in reinforcement learning are described in this section.
}%
{%
  全体のシステム構成, 強化学習におけるAction, State, Rewardの定義を順に述べる.
}%

\subsection{System Architecture}
\begin{figure}[t]
  \centering
  \includegraphics[width=0.95\columnwidth]{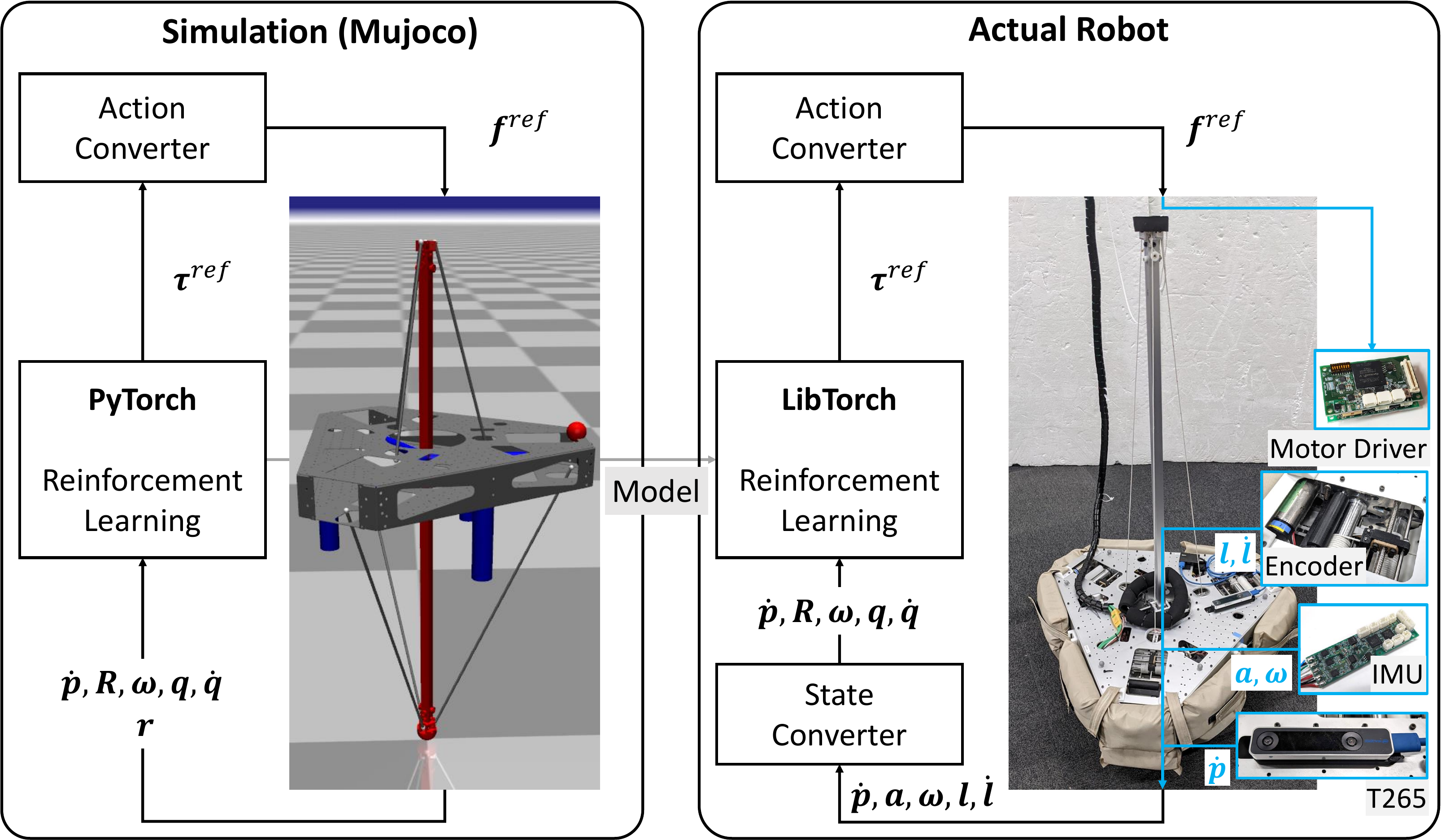}
  \vspace{-1.0ex}
  \caption{System architecture of reinforcement learning for simulation and the actual robot of RAMIEL.}
  \vspace{-3.0ex}
  \label{figure:system}
\end{figure}

\switchlanguage%
{%
  The system configuration of this study is shown in \figref{figure:system}.
  First, the simulation is conducted using Mujoco \cite{todorov2012mujoco}.
  Reinforcement learning is performed using the Proximal Policy Optimization (PPO) \cite{schulman2017ppo} of Stable Baselines3 \cite{stable-baselines3}.
  Here, the simulations show that the body velocity $\bm{\dot{p}}$ ($\in R^{3}$), body rotation matrix $\bm{R}$ ($\in R^{3\times3}$), body angular velocity $\bm{\omega}$ ($\in R^{3}$), joint angle $\bm{q}$ ($\in R^{3}$), and joint velocity $\bm{\dot{q}}$ ($\in R^{3}$) are obtained as the states.
  We also compute the reward $\bm{r}$.
  The action in reinforcement learning is the target joint torque $\bm{\tau}^{ref}$ ($\in R^{3}$).
  This is converted to the target wire tension $\bm{f}^{ref}$ ($\in R^{6}$) by Action Converter of \secref{subsec:action} and sent to the robot.

  On the other hand, in the actual robot, the PyTorch model obtained from Stable Baselines3 is converted for C++, and only inference is performed using LibTorch.
  The wire length $\bm{l}$ ($\in R^{6}$) and the wire velocity $\bm{\dot{l}}$ ($\in R^{6}$) are obtained from the motor encoder.
  In addition, the body acceleration $\bm{\ddot{p}}$ ($\in R^{3}$) and the body angular velocity $\bm{\omega}$ ($\in R^{3}$) are obtained from the inertial sensor, and the body velocity $\bm{\dot{p}}$ ($\in R^{3}$) is obtained from Realsense T265 which performs visual odometry.
  These are converted by State Converter of \secref{subsec:state} into $\{\bm{\dot{p}}, \bm{R}, \bm{\omega}, \bm{q}, \bm{\dot{q}}\}$ which are the same state inputs as in the simulation.
  From these, $\bm{\tau}^{ref}$ is calculated using the trained model and converted to $\bm{f}^{ref}$ by Action Converter exactly as in the simulation, and used as the control input to the motor driver.
  The $\bm{f}^{ref}$ is converted to current and the motor is driven by current control.
  Since this is a quasi direct-drive mechanism, it is possible to achieve the target tension precisely.

  The actual behavior $\bm{a}$ output by the reinforcement learning model is not the direct output of $\bm{\tau}^{ref}$, but the converted value within the range of $[-1, 1]$.
  In other words, when the minimum and maximum values of $\bm{\tau}$ are $\bm{\tau}^{\{min, max\}}$, $\tau^{ref} = \tau^{min} + (\tau^{max}-\tau^{min})(a+1)/2$ for each joint axis.
  The state $\bm{s}$ actually input to the reinforcement learning model is obtained by processing $\{\bm{\dot{p}}, \bm{R}, \bm{\omega}, \bm{q}, \bm{\dot{q}}\}$, which is described in \secref{subsec:state}.
  In this study, $\bm{\tau}^{min}=\{-50, -50, -320\}$, $\bm{\tau}^{max}=\{50, 50, 90\}$ (referring to roll, pitch and linear DOFs in that order).
  Note that this control loop is executed at 100 Hz.
}%
{%
  本研究のシステム構成を\figref{figure:system}に示す.
  まず, シミュレーションはMujoco \cite{todorov2012mujoco}を使い構成している.
  Stable Baselines3 \cite{stable-baselines3}のProximal Policy Optimization (PPO) \cite{schulman2017ppo}を用いて強化学習を実行する.
  この際, シミュレーションからは本体速度$\bm{\dot{p}}$ ($\in R^{3}$), 本体の回転行列$\bm{R}$ ($\in R^{3\times3}$) , 本体の角速度$\bm{\omega}$ ($\in R^{3}$), 関節角度$\bm{q}$ ($\in R^{3}$), 関節速度$\bm{\dot{q}}$ ($\in R^{3}$)を状態として得る.
  また, 報酬$\bm{r}$を計算する.
  強化学習における行動は指令関節トルク$\bm{\tau}^{ref}$ ($\in R^{3}$)である.
  これは, \secref{subsec:action}のAction Converterにより指令ワイヤ張力$\bm{f}^{ref}$ ($\in R^{6}$)に変換され, ロボットに送られる.

  一方, 実機ではStable Baselines3から得られたPyTorchのモデルをC++用に変換し, LibTorchを用いることで推論のみを行う.
  ロボット実機では, モータのエンコーダからワイヤ長$\bm{l}$ ($\in R^{6}$)とワイヤ速度$\bm{\dot{l}}$ ($\in R^{6}$)が得られる.
  また, 慣性センサから本体加速度$\bm{\ddot{p}}$ ($\in R^{3}$), 本体角速度$\bm{\omega}$ ($\in R^{3}$), Visual Odometryを行うRealsense T265からは本体速度$\bm{\dot{p}}$ ($\in R^{3}$)が得られる.
  これらは, \secref{subsec:state}のState Converterによって, シミュレーションと同様の入力である$\{\bm{\dot{p}}, \bm{R}, \bm{\omega}, \bm{q}, \bm{\dot{q}}\}$に変換される.
  これらから, 訓練されたモデルを用いて$\bm{\tau}^{ref}$を計算し, シミュレーションと全く同様のAction Converterにより$\bm{f}^{ref}$に変換して, モータドライバへの制御入力としている.
  $\bm{f}^{ref}$を電流へと変換し, 電流制御によりモータを駆動する.
  準ダイレクトドライブ機構であるため, 正確に指令した張力を実現することが可能である.

  なお, 実際に強化学習モデルが出力する行動$\bm{a}$は, $\bm{\tau}^{ref}$を直接用いるのではなく, $[-1, 1]$の範囲に含まれる値に変換した値としている.
  つまり, $\bm{\tau}$の最小値・最大値を$\bm{\tau}^{\{min, max\}}$としたとき, それぞれの関節軸に対して$\tau^{ref} = \tau^{min} + (\tau^{max}-\tau^{min})(a+1)/2$が成り立つ.
  また, 実際に強化学習モデルに入力する状態$\bm{s}$は, 得られた$\{\bm{\dot{p}}, \bm{R}, \bm{\omega}, \bm{q}, \bm{\dot{q}}\}$を加工して用いており, これについては\secref{subsec:state}で述べる.
  本研究では$\bm{\tau}^{min}=\{-50, -50, -320\}$, $\bm{\tau}^{max}=\{50, 50, 90\}$とする(順にロール, ピッチ, 直動自由度を指す).
  なお, 本制御ループは100 Hzで実行されている.

}%

\subsection{Definition of Action} \label{subsec:action}
\switchlanguage%
{%
  In this study, although the control input of the actual robot is the wire tension $\bm{f}^{ref}$, $\bm{f}^{ref}$ is not used as the direct control input because of the increase in learning time due to the increase of search range and the increase in internal force by wire antagonism.
  Instead, the joint torque $\bm{\tau}^{ref}$ is used as the control input in reinforcement learning, and $\bm{\tau}^{ref}$ is converted to $\bm{f}^{ref}$ in exactly the same way in simulation and in the actual robot (Action Converter).
  By solving the following quadratic programming, we calculate $\bm{f}^{ref}$ that satisfies $\bm{\tau}^{ref}$ while minimizing the internal forces:
  \begin{align}
    \underset{\bm{f}^{ref}}{\textrm{minimize}}&\;\;\;\;\;\;\;\;\;\;\;\;\;\;\;\;\;\;\;||\bm{f}^{ref}||^{2}\label{eq:balance}\\
    \textrm{subject to}&\;\;\;\;\;\;\;\;\;\;\;\; \bm{\tau}^{ref} = -G^{T}(\bm{q})\bm{f}^{ref}\nonumber\\
    &\;\;\;\;\;\;\;\;\;\;\;\; \bm{f}^{ref} \geq \bm{f}^{min} \nonumber
  \end{align}
  where $\bm{G}$ is the muscle Jacobian of the wire at the joint angle $\bm{q}$ (which can be calculated from the geometric model), $\bm{f}^{min}$ is the minimum muscle tension, and $||\cdot||$ is the L2 norm.
  In this study, we set $\bm{f}^{min}=8$ [N].
}%
{%
  本研究では, 実際のロボットの制御入力はワイヤ張力$\bm{f}^{ref}$であるものの, これを直接強化学習のActionとした場合, ワイヤの拮抗による内力上昇や探索範囲増大による学習時間の増大の観点から, $\bm{f}^{ref}$を直接制御入力としては用いない.
  その代わりに, 関節トルク$\bm{\tau}^{ref}$を強化学習における制御入力として, シミュレーションと実機で全く同様の形で$\bm{\tau}^{ref}$を$\bm{f}^{ref}$に変換している(Action Converter).
  以下の二次計画法を解くことで, 内力を最小化しつつ$\bm{\tau}^{ref}$を満たす$\bm{f}^{ref}$を計算する.
  \begin{align}
    \underset{\bm{f}^{ref}}{\textrm{minimize}}&\;\;\;\;\;\;\;\;\;\;\;\;\;\;\;\;\;\;\;||\bm{f}^{ref}||^{2}\label{eq:balance}\\
    \textrm{subject to}&\;\;\;\;\;\;\;\;\;\;\;\; \bm{\tau}^{ref} = -G^{T}(\bm{q})\bm{f}^{ref}\nonumber\\
    &\;\;\;\;\;\;\;\;\;\;\;\; \bm{f}^{ref} \geq \bm{f}^{min} \nonumber
  \end{align}
  ここで, $\bm{G}$は関節角度$\bm{q}$におけるワイヤの筋長ヤコビアン(幾何モデルから計算することが可能), $\bm{f}^{min}$は最小筋張力, $||\cdot||$はL2ノルムを表す.
  本研究では$\bm{f}^{min}=8$ [N]としている.
}%

\subsection{Definition of State} \label{subsec:state}
\switchlanguage%
{%
  We convert $\{\bm{\dot{p}}, \bm{\ddot{p}}, \bm{\omega}, \bm{l}, \bm{\dot{l}}\}$ obtained in the actual robot to the same $\{\bm{\dot{p}}, \bm{R}, \bm{\omega}, \bm{q}, \bm{\dot{q}}\}$ as in the simulation (State Converter).
  First, by using $\bm{\ddot{p}}$ and $\bm{\omega}$ obtained from IMU, the rotation matrix $\bm{R}$ of the main body is calculated by Madgwick Filter \cite{madgwick2010filter}.
  Next, since the joint angles $\bm{q}$ and joint velocities $\bm{\dot{q}}$ of the actual robot cannot be obtained directly, we estimate them from $\bm{l}$ and $\bm{\dot{l}}$.
  The following Extended Kalman Filter (EKF) is performed with \equref{eq:ekf-predict} as the prediction function and \equref{eq:ekf-observe} as the observation function:
  \begin{align}
    \bm{x}_{t+1} &= \begin{pmatrix}\bm{I} & \bm{I}dt\\ \bm{O} & \bm{I}\end{pmatrix}\bm{x}_{t} + \bm{w}\label{eq:ekf-predict}\\
    \bm{y}_{t+1} &= \bm{h}(\bm{x}_{t}) + \bm{v}\label{eq:ekf-observe}\\
    \bm{h}(\bm{x}_{t}) &= \begin{pmatrix}\bm{g}(\bm{q}_{t})\\\bm{G}(\bm{q}_{t})\bm{\dot{q}}_{t}\end{pmatrix}, \bm{x} = \begin{pmatrix}\bm{q}\\\bm{\dot{q}}\end{pmatrix}, \bm{y} = \begin{pmatrix}\bm{l}\\\bm{\dot{l}}\end{pmatrix}\nonumber
  \end{align}
  where $\bm{I}$ is the identity matrix, $\bm{O}$ is the zero matrix, $\{\bm{w}, \bm{v}\}$ is the Gaussian noise, and $\bm{g}(\bm{q})$ is the wire length at the joint angle $\bm{q}$ ($\bm{G}(\bm{q})$ is its derivative).

  We construct the state input $\bm{s}$ for reinforcement learning using $\{\bm{\dot{p}}, \bm{R}, \bm{\omega}, \bm{q}, \bm{\dot{q}}\}$ obtained so far.
  First, the simplest configuration is the one in which $\{\bm{\dot{p}}_{t}, \bm{R}^{quat}_{t}, \bm{\omega}_{t}, \bm{q}_{t}, \bm{\dot{q}}_{t}\}$ are the state values (Ours-1).
  Note that $\bm{R}^{quat}$ is the value obtained by transforming the rotation matrix $\bm{R}$ into quaternion ($\in R^{4}$).
  Next, we construct a state value that takes into account the problem of this study, i.e., the oscillation of the wire length due to its stretching or loosening, and the resulting oscillation of the joint angle estimation (Ours-2).
  We use the characteristics that $\bm{q}$ is less sensitive to the vibration than $\bm{\dot{q}}$.
  Instead of using $\dot{\bm{q}}$ directly, we set $N_{prev}$, which indicates how many previous time steps to be given, and use $\bm{q}$ of the previous steps as the state value.
  That is, instead of $\dot{\bm{q}}_{t}$, we use $\{\bm{q}_{t-1}, \bm{q}_{t-2}, \cdots, \bm{q}_{t-N_{prev}}\}$ as state values.
  In addition to these state values, both Ours-1 and Ours-2 use $\{\bm{\tau}^{ref}_{t-1}, \bm{\tau}^{ref}_{t-2}, \cdots, \bm{\tau}^{ref}_{t-N_{prev}}\}$ as the state value.
  Note that we set $N_{prev}=6$ in this study.
  Ours-1 and Ours-2 are compared in the experiment.
}%
{%
  実機で得られた$\{\bm{\dot{p}}, \bm{\ddot{p}}, \bm{\omega}, \bm{l}, \bm{\dot{l}}\}$をシミュレーションと同様の$\{\bm{\dot{p}}, \bm{R}, \bm{\omega}, \bm{q}, \bm{\dot{q}}\}$に変換する(State Converter).
  まず, IMUから得られた$\bm{\ddot{p}}$と$\bm{\omega}$を用いて, Madgwick Filter \cite{madgwick2010filter}により, 本体の回転行列$\bm{R}$を計算する.
  次に, 実機の関節角度$\bm{q}$と関節速度$\bm{\dot{q}}$は直接得られないため, $\bm{l}$と$\bm{\dot{l}}$からこれを推定する.
  以下の\equref{eq:ekf-predict}を遷移関数, \equref{eq:ekf-observe}を観測関数として, 拡張カルマンフィルタを実行する.
  \begin{align}
    \bm{x}_{t+1} &= \begin{pmatrix}\bm{I} & \bm{I}dt\\ \bm{O} & \bm{I}\end{pmatrix}\bm{x}_{t} + \bm{w}\label{eq:ekf-predict}\\
    \bm{y}_{t+1} &= \bm{h}(\bm{x}_{t}) + \bm{v}\label{eq:ekf-observe}\\
    \bm{h}(\bm{x}_{t}) &= \begin{pmatrix}\bm{g}(\bm{q}_{t})\\\bm{G}(\bm{q}_{t})\bm{\dot{q}}_{t}\end{pmatrix}, \bm{x} = \begin{pmatrix}\bm{q}\\\bm{\dot{q}}\end{pmatrix}, \bm{y} = \begin{pmatrix}\bm{l}\\\bm{\dot{l}}\end{pmatrix}\nonumber
  \end{align}
  ここで, $\bm{I}$は単位行列, $\bm{O}$はゼロ行列, $\{\bm{w}, \bm{v}\}$はガウス雑音, $\bm{g}(\bm{q})$は関節角度$\bm{q}$におけるワイヤ長($\bm{G}(\bm{q})$はこの微分)を表す.

  これまでに得られた$\{\bm{\dot{p}}, \bm{R}, \bm{\omega}, \bm{q}, \bm{\dot{q}}\}$を用いて強化学習の状態入力$\bm{s}$を構成する.
  まず, 最も単純な構成はこれらを並べた構成で, $\{\bm{\dot{p}}_{t}, \bm{R}^{quat}_{t}, \bm{\omega}_{t}, \bm{q}_{t}, \bm{\dot{q}}_{t}\}$を状態値とする場合である(Ours-1).
  なお, ここで$\bm{R}^{quat}$は回転行列$\bm{R}$をQuaternion ($\in R^{4}$)に変換した値とする.
  次に, 本研究の問題点である, ワイヤの伸びや緩みに伴うワイヤの振動, その結果としての関節角度推定値の振動を考慮した状態値を構成する(Ours-2).
  $\bm{q}$は$\bm{\dot{q}}$はに比べて振動の影響を受けにくいという性質を用いる.
  $\dot{\bm{q}}$は直接用いず, いくつ前のタイムステップの情報まで与えるかを表す$N_{prev}$を設定し, その数だけ前ステップの$\bm{q}$を状態値として用いる.
  つまり, $\dot{\bm{q}}_{t}$の代わりに$\{\bm{q}_{t-1}, \bm{q}_{t-2}, \cdots, \bm{q}_{t-N_{prev}}\}$を状態値に用いる.
  これらの状態値に加え, Ours-1とOurs-2の両者ともこれまでに送った指令値の情報として$\{\bm{\tau}^{ref}_{t-1}, \bm{\tau}^{ref}_{t-2}, \cdots, \bm{\tau}^{ref}_{t-N_{prev}}\}$を状態値に加える.
  なお, 本研究では$N_{prev}=6$とする.
  このOurs-1とOurs-2は実験において比較する.
}%

\subsection{Definition of Reward} \label{subsec:reward}
\switchlanguage%
{%
  In this study, we design the following rewards to generate jumping behaviors.
  First, the following reward $r_{jump}$ is given for the height of the body in order to generate continuous jumps from the landing state:
  \begin{align}
    r_{jump} = \begin{cases} -1 & (p_{z}>1.1)\\ p^{2}_{z} & (otherwise)\end{cases} \label{eq:reward-jump}
  \end{align}
  where $p_{z}$ is the height of the body.
  The reward is proportional to the square of the height of the body, and at the same time, a penalty is given when the height exceeds 1.1 m in order to prevent jumping too high.
  Although $\bm{p}_{z}$ is obtained from Realsense T265, we do not use it as a state value because it may be significantly off due to noise.

  Next, the following reward $r_{keep}$ is given to the translational velocity of the body in the $xy$ direction and the rotational velocity around the $z$-axis, so that the body makes a jump in-situ:
  \begin{align}
    r_{keep} = - c(\dot{p}^{2}_{x} + \dot{p}^{2}_{y} + \omega^{2}_{z}) \label{eq:reward-keep}
  \end{align}
  Note that $c$ is a variable that monotonically increases from 0 to 1 according to the process of learning.

  Next, the following reward $r_{horizon}$ is given to keep the body and the leg horizontal and suppress tilting:
  \begin{align}
    r_{horizon} = - (1.0 - (\bm{R}\cdot\bm{e}_{z})\cdot\bm{e}_{z}) - (1.0 - (\bm{R}_{leg}\cdot\bm{e}_{z})\cdot\bm{e}_{z}) \label{eq:reward-horizon}
  \end{align}
  where $\bm{R}_{leg}$ is the rotation matrix of the leg in the world coordinate and $\bm{e}_{z}=\begin{pmatrix}0 & 0 & 1\end{pmatrix}^{T}$.
  Each for the body and leg, the current vector along the $z$-axis ($\bm{R}\cdot\bm{e}_{z}$) is taken, and the degree of tilt is calculated from the inner product of this vector and the $z$-axis in the world coordinate.
  When both the body and leg are horizontal, $r_{horizon}$ is zero, and the value increases negatively with the degree of tilt.

  Next, the following reward $r_{ctrl}$ is given in order to suppress the increase of control input:
  \begin{align}
    r_{ctrl} = - ||\bm{w}_{a}\cdot\bm{a}||^{2} \label{eq:reward-ctrl}
  \end{align}
  where $\bm{w}_{a}=\begin{pmatrix}3.0&3.0&0.3\end{pmatrix}^{T}$.
  Since a large force is required in the $z$ direction, the weight for the slide joint is set to be small.

  Next, the following reward $r_{contact}$ is given, which reduces the time when the leg and the landing legs of the body make contact with the floor, so that the leg is correctly lifted and the jumping motion is performed:
  \begin{align}
    r^{leg}_{contact} &= \begin{cases} -0.3 - 3000p^{2}_{leg/tip, z} & (p_{leg/tip, z} < 0) \\ 0 & (otherwise) \end{cases}\\
    r^{land}_{contact} &= \begin{cases} -0.3 & (p_{land/tip, z} < 0) \\ 0 & (otherwise) \end{cases}\\
    r_{contact} &= r^{leg}_{contact} + r^{land}_{contact} \label{eq:reward-contact}
  \end{align}
  where $p_{leg/tip, z}$ is the height of the tip of the leg with the floor height set to 0, and $p_{land/tip, z}$ is the height of the tip of the landing legs of the body.
  The robot is encouraged to jump from the position where the leg and the body have landed on the floor.
  In addition, when the speed of the leg tip at landing is high, the leg tip may roll into the floor due to the simulation environment.
  In order to prevent this phenomenon and to make the jump while landing softly on the ground, a penalty is given for the distance the foot tip sinks.

  Finally, the following reward $r_{range}$ is given to constrain the range of joint angles:
  \begin{align}
    r_{range, i} &= \begin{cases}
      -({q}^{min}_{i}+{q}^{thre}_{i}-{q}_{i})^{2} & ({q}_{i} < {q}^{min}_{i}+{q}^{thre}_{i})\\
      -({q}^{max}_{i}-{q}^{thre}_{i}-{q}_{i})^{2} & ({q}_{i} > {q}^{max}_{i}-{q}^{thre}_{i})\\
    \end{cases}\label{eq:reward-range-i}\\
    r_{range} &= 10r_{range, r} + 10r_{range, p} + 50r_{range, s} \label{eq:reward-range}
  \end{align}
  where ${q}^{\{min, max, thre\}}_{i}$ denotes the minimum, maximum and threshold values to constrain $q$, respectively, and $i$ is one of $\{r, p, s\}$ representing the joint DOFs $\{roll, pitch, slide\}$.
  We set $\bm{q}=\begin{pmatrix}q_{r} & q_{p} & q_{s}\end{pmatrix}^{T}$.
  In other words, a penalty is given when the joint angle limit is approached beyond a certain threshold value.
  In this study, we set ${q}^{min}_{\{r, p, s\}}=\{-0.8, -0.8, 0.1\}$, ${q}^{max}_{\{r, p, s\}}=\{0.8, 0.8, 0.926\}$, and ${q}^{thre}_{\{r, p, s\}}=\{0.4, 0.4, 0.15\}$.

  The following reward $r$, which is the sum of the rewards described so far, is used for learning:
  \begin{align}
    r = &r_{jump} + r_{keep} + r_{horizon} + r_{ctrl}\nonumber\\
        & + r_{contact} + r_{range} + 0.1 \label{eq:reward-total}
  \end{align}
  The last 0.1 is the survival reward given just for surviving without reaching the termination condition of the simulation described in \secref{subsec:parameters}.
}%
{%
  本研究では以下のような報酬を設計してジャンプ動作を生成した.
  まず, 着地している状態から連続ジャンプを行うために, 本体の高さについて以下の報酬$r_{jump}$を与える.
  \begin{align}
    r_{jump} = \begin{cases} -1 & (p_{z}>1.1)\\ p^{2}_{z} & (otherwise)\end{cases} \label{eq:reward-jump}
  \end{align}
  ここで, $p_{z}$は本体の高さを表す.
  本体の高さの二乗に比例した報酬を与えると同時に, 高過ぎるジャンプを抑制するため, 高さが1.1 mを超えた際はペナルティを与えている.
  $\bm{p}_{z}$はRealsense T265から得られるが, ノイズによって大きくズレることがあるため, 状態値としては用いていない.

  次に, その場でジャンプを行うように, 本体の$xy$方向の並進速度とz軸回りの回転速度に対して以下の報酬$r_{keep}$を与える.
  \begin{align}
    r_{keep} = - c(\dot{p}^{2}_{x} + \dot{p}^{2}_{y} + \omega^{2}_{z}) \label{eq:reward-keep}
  \end{align}
  なお, $c$は学習のタイムステップに従って0から1まで単調増加する変数である.

  次に, 本体と脚の傾きを抑えるために, これらを水平に保つ以下の報酬$r_{horizon}$を与える.
  \begin{align}
    r_{horizon} = - (1.0 - (\bm{R}\cdot\bm{e}_{z})\cdot\bm{e}_{z}) - (1.0 - (\bm{R}_{leg}\cdot\bm{e}_{z})\cdot\bm{e}_{z}) \label{eq:reward-horizon}
  \end{align}
  ここで, $\bm{R}_{leg}$は脚の世界座標系における回転行列, $\bm{e}_{z}=\begin{pmatrix}0 & 0 & 1\end{pmatrix}^{T}$である.
  本体と脚それぞれに対して, 現在のz軸方向のベクトル($\bm{R}\cdot\bm{e}_{z}$)を取り出し, これと世界座標系におけるz軸の内積から傾きの程度を算出している.
  本体・脚ともに水平であるとき, $r_{horizon}$は0となり, 傾くにつれて負に大きくなる.

  次に, 制御入力の上昇を抑えるために, 以下の報酬$r_{ctrl}$を与える.
  \begin{align}
    r_{ctrl} = - ||\bm{w}_{a}\cdot\bm{a}||^{2} \label{eq:reward-ctrl}
  \end{align}
  ここで, $\bm{w}_{a}=\begin{pmatrix}3.0&3.0&0.3\end{pmatrix}^{T}$である.
  $z$方向は大きな力が必要であるため, 重みを小さく設定している.

  次に, 正しく脚を持ち上げジャンプ動作を行うように, 脚と本体の着地脚が床に接触する時間を減らす以下の報酬$r_{contact}$を与える.
  \begin{align}
    r^{leg}_{contact} &= \begin{cases} -0.3 - 3000p^{2}_{leg/tip, z} & (p_{leg/tip, z} < 0) \\ 0 & (otherwise) \end{cases}\\
    r^{land}_{contact} &= \begin{cases} -0.3 & (p_{land/tip, z} < 0) \\ 0 & (otherwise) \end{cases}\\
    r_{contact} &= r^{leg}_{contact} + r^{land}_{contact} \label{eq:reward-contact}
  \end{align}
  ここで, $p_{leg/tip, z}$は床の高さを0とした脚先の高さ, $p_{land/tip, z}$は本体についた着地脚の先端の高さを表す.
  脚と本体が床に着地した状態からジャンプすることを促す.
  また, 着地時の脚先速度が高いと, 脚先が床にめり込む現象が起こる.
  これを防ぎ, 柔らかく地面に着地しながらジャンプを行うため, めり込んだ距離に対してペナルティを与えている.

  最後に, 関節角度範囲に対して制約をかけるため, 以下の報酬$r_{range}$を与える.
  \begin{align}
    r_{range, i} &= \begin{cases}
      -({q}^{min}_{i}+{q}^{thre}_{i}-{q}_{i})^{2} & ({q}_{i} < {q}^{min}_{i}+{q}^{thre}_{i})\\
      -({q}^{max}_{i}-{q}^{thre}_{i}-{q}_{i})^{2} & ({q}_{i} > {q}^{max}_{i}-{q}^{thre}_{i})\\
    \end{cases}\label{eq:reward-range-i}\\
    r_{range} &= 10r_{range, r} + 10r_{range, p} + 50r_{range, s} \label{eq:reward-range}
  \end{align}
  ここで, ${q}^{\{min, max, thre\}}_{i}$はそれぞれ$q$の最小値・最大値・制約をかける閾値を表し, $i$には関節自由度$\{roll, pitch, slide\}$を表す$\{r, p, s\}$のいずれかが入る.
  $\bm{q}=\begin{pmatrix}q_{r} & q_{p} & q_{s}\end{pmatrix}^{T}$としている.
  つまり, ある閾値を超えて関節角度限界に近づいた場合に, ペナルティを与えている.
  なお, 本研究では${q}^{min}_{\{r, p, s\}}=\{-0.8, -0.8, 0.1\}$, ${q}^{max}_{\{r, p, s\}}=\{0.8, 0.8, 0.926\}$, ${q}^{thre}_{\{r, p, s\}}=\{0.4, 0.4, 0.15\}$とした.

  これまでに述べた報酬を足した合わせた以下の報酬$r$を用いて学習を行う.
  \begin{align}
    r = r_{jump} + r_{keep} + r_{horizon} + r_{ctrl} + r_{contact} + r_{range} + 0.1 \label{eq:reward-total}
  \end{align}
  最後の0.1は, \secref{subsec:parameters}で述べるシミュレーションの終了条件に達さずに生存するだけで与える生存報酬である.
}%

\subsection{Other Parameters for Reinforcement Learning} \label{subsec:parameters}
\switchlanguage%
{%
  First, we describe the setup of reinforcement learning algorithm PPO.
  In this study, we set the total number of learning steps to 1.0E+7, the number of parallel environments to 6, the number of steps used for training at a time to 6$\times$2048, the batch size to 1024, the number of epochs to 10, and other default values to those of \cite{schulman2017ppo}.
  The network structure uses fully-connected layers with [256, 128] number of units in hidden layers.
  When the number of steps of the episodes exceeds 1.0E+4, the joint angle $\bm{q}$ deviates from $\bm{q}^{\{min, max\}}$ set in \equref{eq:reward-range-i}, or $(\bm{R}_{leg}\cdot\bm{e}_{z})\cdot\bm{e}_{z}$ set in \equref{eq:reward-keep} exceeds 0.5, the episode is terminated.

  Next, we discuss the introduction of noise into the state and action.
  The actual robot and the simulation are very different, so, by imitating the noise characteristics and reflecting them in the simulation, the reinforcement learning results in the simulation can be smoothly transferred to the actual robot.
  For $\bm{q}$ and $\bm{R}^{quat}$, we add noise of $cN(0, 0.05)$ to the state values (where $N(\mu, \sigma)$ is the Gaussian noise with mean $\mu$ and variance $\sigma$).
  For $\bm{\dot{q}}$ and $\bm{\dot{p}}$, we add $c||\bm{\dot{q}}||N(0, 0.05)$ and $c|||\bm{\dot{p}}||N(0, 0.1)$ noise to the state values, respectively.
  For actions, we multiply $k_f=\textrm{min}(0.8 + N(0, 0.1), 1.0)$ by $f^{ref}$ for each wire tension calculated by the Action Converter, giving the loss due to random friction.
  Also, this $k_f$ is varied step by step as $k_f \gets k_f + cN(0, 0.01)$.

  Finally, we discuss other types of noises related to external forces and initial posture.
  For the initial posture, we add noise of $cN(0, 0.03)$ [rad] to the joint angles of roll and pitch $\{q^{init}_{\{r, p\}}\}$ at the initial position on the ground.
  As for the external forces, we apply forces of $cN(0, 10)$ [N] to the translational direction of the body, respectively.
}%
{%
  まず, 強化学習であるPPOの設定について述べる.
  本研究では, 学習の合計ステップ数を1.0E+7, 並列環境数を6, 一度の学習に用いるステップ数を6$\times$2048, バッチサイズを1024, エポック数を10, それ以外はデフォルトの値とした.
  ネットワーク構造は, 隠れ層数が[256, 128]の全結合層を使用している.
  また, そのエピソードのステップ数が1.0E+4を超える, 関節角度$\bm{q}$が\equref{eq:reward-range-i}で設定した$\bm{q}^{\{min, max\}}$を逸脱する, または\equref{eq:reward-keep}で設定した$(\bm{R}_{leg}\cdot\bm{e}_{z})\cdot\bm{e}_{z}$が0.5を超えた時, そのエピソードを終了する.

  次に, 状態の観測と行動に対するノイズの導入について述べる.
  実機とシミュレーションは大きく異なり, 特にそのノイズ特性を真似てシミュレーション側に反映させることで, シミュレーションにおける強化学習結果をスムーズに実機に転移することができる.
  $\bm{q}$と$\bm{R}^{quat}$に対して, $cN(0, 0.05)$のノイズを加えて状態値とする(ここで, $N(\mu, \sigma)$は平均$\mu$, 分散$\sigma$のガウスノイズとする).
  $\bm{\dot{q}}$と$\bm{\dot{p}}$に対して$c||\bm{\dot{q}}||N(0, 0.05)$, $c||\bm{\dot{p}}||N(0, 0.1)$のノイズをそれぞれ加えて状態値とする.
  行動については, Action Converterにより計算されたそれぞれのワイヤの$f^{ref}$に対して, $k_f=\textrm{min}(0.8 + N(0, 0.1), 1.0)$をかけ, ランダムな摩擦による損失を与える.
  また, この$k_f$はステップごとに$k_f \gets k_f + cN(0, 0.01)$のように変化させる.

  最後に, その他の外力や初期姿勢等に関するノイズについて述べる.
  初期姿勢については, 地面に接地した初期位置におけるロールとピッチの関節角度$\{q^{init}_{\{r, p\}}\}$に対して$cN(0, 0.03)$ [rad]のノイズを加える.
  外力については, 本体の並進方向に対してそれぞれ$cN(0, 10)$ [N]の力を加えている.
}%

\section{Experiments} \label{sec:experiment}

\subsection{Simulation Experiment}
\begin{figure}[t]
  \centering
  \includegraphics[width=0.8\columnwidth]{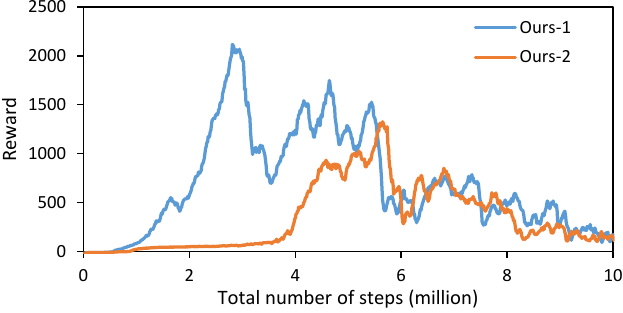}
  \vspace{-1.0ex}
  \caption{Transition of reward during training.}
  \label{figure:training}
\end{figure}

\begin{figure}[t]
  \centering
  \includegraphics[width=0.9\columnwidth]{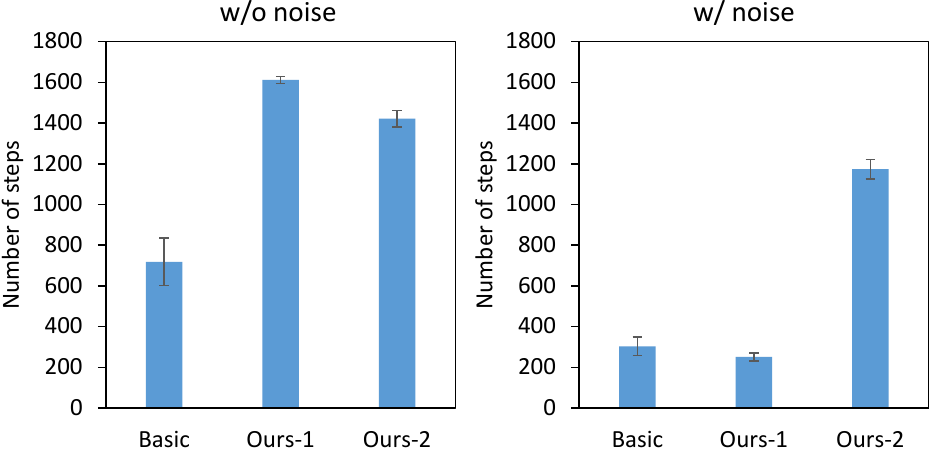}
  \vspace{-1.0ex}
  \caption{Result of simulation experiments for Basic, Ours-1, and Ours-2 controllers.}
  \vspace{-3.0ex}
  \label{figure:sim-exp}
\end{figure}

\switchlanguage%
{%
  First, the training results for Ours-1 and Ours-2 are shown in \figref{figure:training}.
  Since Ours-1 directly uses joint velocities as states, the training progress is faster than that of Ours-2, which has to estimate joint velocities from joint angle sequences.
  On the other hand, the reward values become similar as the learning progresses.
  In addition, some penalties and noise increase as the learning progresses, indicating that the rewards are gradually decreasing.

  Next, the results of the general controller (Basic) of \cite{raibert1984onelegged} and our methods Ours-1 and Ours-2 are shown in \figref{figure:sim-exp}.
  We compared the mean and variance of the number of steps that were followed by a jump for each of five trials in both the noiseless environment and in the environment where $N(0, 0.001)$ noise is added to the muscle length.
  In the noiseless environment, Ours-1 and Ours-2 outperform Basic.
  On the other hand, in the noisy environment, the performance of Basic and Ours-1 deteriorates significantly, and Ours-2 performs the best.
  The main reason for this is considered to be that the control methods of Basic and Ours-1 are dependent on the joint velocity.
}%
{%
  まず, Ours-1とOurs-2における訓練結果を\figref{figure:training}に示す.
  Ours-1は直接関節速度を状態として用いるため, 関節角度列から関節速度を推定しなければならないOurs-2に比べて学習の進みが速い.
  一方で, 学習が進むにつれてその報酬値は同様の値を示すようになる.
  また, 一部のペナルティやノイズは学習が進むごとに増加するため, 報酬は徐々に下がっていることがわかる.

  次に, \cite{raibert1984onelegged}の制御器(Basic), Ours-1, Ours-2を用いた本手法の結果を\figref{figure:sim-exp}に示す.
  ノイズのない環境と, 筋長に$N(0, 0.001)$のノイズを加えた環境の両者において, 5回ずつ動作を行い跳躍が続いたステップ数の平均と分散を比較した.
  ノイズのない環境では, Basicに比べ, Ours-1とOurs-2が大きく性能を上回る結果となった.
  一方で, ノイズのある環境では, BasicとOurs-1の性能は大きく劣化し, Ours-2が最も良い結果となった.
  これは, BasicとOurs-1は関節速度に依存した制御手法となっていることが大きな要因だと考えられる.
}%

\begin{figure*}[t]
  \centering
  \includegraphics[width=2.0\columnwidth]{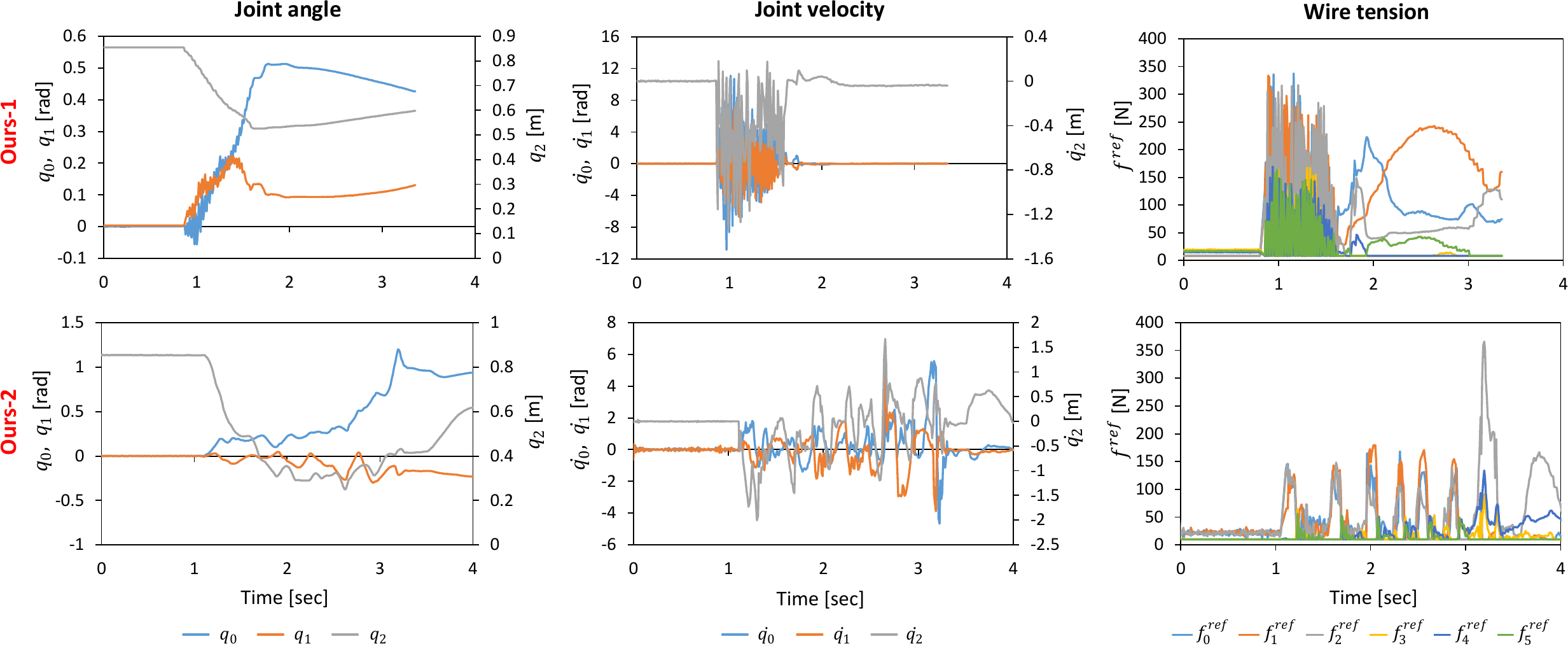}
  \caption{Result of the actual robot experiments for Ours-1 and Ours-2 controllers.}
  \label{figure:act-exp}
\end{figure*}

\subsection{Actual Robot Experiment}
\switchlanguage%
{%
  The results of applying Ours-1 and Ours-2 to the actual robot are shown in \figref{figure:act-exp}.
  For the case of Ours-1, it can be seen that the wire length oscillates as wire tension is generated, and the estimated joint velocities accordingly oscillate and diverge significantly.
  As a result, the target wire tension output from the model also oscillates significantly, and the robot falls down without being able to jump even once.

  On the other hand, for Ours-2, although the joint velocity oscillates slightly, the wire tension exerts appropriate force without being affected by the oscillation, and as a result, the robot is able to realize stable motion without divergence.
  In the present movement, the robot successfully jumped four times in succession.
  Because the body shifted in the translational direction, tension was applied to the apparatus suspending RAMIEL from above at the fifth jump.
}%
{%
  Ours-1とOurs-2を実機に適用した結果を\figref{figure:act-exp}に示す.
  Ours-1についてはワイヤ張力の発生に伴いワイヤ長が振動し, それに伴って関節速度の推定値が大きく振動, 発散していることがわかる.
  これにより, モデルから出力される指令ワイヤ張力も大きく振動し, 一度もジャンプできずに倒れてしまった.

  一方で, Ours-2については関節速度は多少振動しているものの, ワイヤ張力がそれに影響を受けずに適切な力を発揮するため, 結果的に発散が起こらず, 安定して動作が実現できている.
  本動作は4回の連続跳躍に成功しており, 体が並進方向にズレていったため, 5回目の跳躍でRAMIELを上から吊る器具に張力がかかってしまった.
}%

\section{Discussion} \label{sec:discussion}
\switchlanguage%
{%
  We discuss the results of this study.
  First, a very dynamic and difficult task such as continuous jumping is made feasible with our reward, state, and action settings in reinforcement learning, with better performance than that of the method described in \cite{raibert1984onelegged}.
  In this case, the joint velocity term is important, and it is shown that the learning progress is accelerated when this term is input as a state.
  On the other hand, since RAMIEL used in this study is driven by antagonistic wires, the effect of the oscillation of joint angles and joint velocities due to wire elongation cannot be ignored.
  In fact, when noise is applied to the wire length, the wire length, joint angle, and joint velocity oscillate accordingly.
  Therefore, while the performance of Ours-1 is better than that of Ours-2 without noise, that of Ours-1 is lower than that of Basic as well as Ours-2 in a noisy environment.
  Since joint angles are less prone to noise than joint velocities, the performance of Ours-2 does not deteriorate significantly in the noisy environment.
  This effect was more pronounced on the actual robot than in simulation.
  The oscillation of the wire length during the generation of wire tension on the actual robot was larger than expected, and Ours-1 could not jump even once.
  The oscillation of the joint velocity and that of the target wire tension alternated, causing the motion to diverge.
  On the other hand, Ours-2 was not significantly affected by the joint velocity oscillation, and the target wire tension was stable, resulting in smoother joint velocity transitions than Ours-1.
  Finally, the robot succeeded in making four consecutive jumps.

  We discuss the problems and future prospects of this study.
  A significant progress has been made in this study in terms of the use of reinforcement learning and its applicability to the dynamic motion of a parallel wire-driven robot with wire elongation.
  On the other hand, we have not yet succeeded in achieving a complete continuous jumping motion.
  We believe that there are three reasons for this.
  (1) It is important to note that the friction of the body, especially the friction term in the joint motion and the loss of the output wire tension, does not match between the simulation and the actual robot.
  This corresponds to Actuator Net in \cite{hwangbo2019anymal}, and Real2Sim using data from the actual robot is likely to be important.
  (2) We believe that it is necessary to add not only Gaussian noise but also steady-state error to the state.
  Depending on the installation of sensors, there are many cases where the noise is not only Gaussian noise but also biased noise.
  It is necessary to consider a noise design that takes such biases into account, as well as a method to reflect the steady-state error of the actual robot in the simulation.
  (3) The delay of observation and action needs to be dealt with.
  In particular, the body velocity output from Realsense T265, the joint angle estimation calculated from the wire length, and the delay from the motor control input to the wire tension generation due to the motor inertia need to be considered.
  We believe that when all of these problems are solved, stable continuous jumping motion will be realized on the actual robot.
}%
{%
  本研究の結果について考察する.
  まず, 連続跳躍のような非常に動的で困難なタスクが, 強化学習における本研究の報酬・状態・行動の設定により可能であり, その性能は\cite{raibert1984onelegged}らの手法を上回る.
  その際, 関節速度項は重要であり, これを状態として入力することで学習の進みが速くなることがわかる.
  一方で, 本研究で扱うRAMIELは拮抗ワイヤ駆動型であるため, ワイヤの伸びによる関節角度・関節速度の振動の影響は無視できない.
  実際, ワイヤ長に対してノイズを入れた場合, それに伴ってワイヤ長, 関節角度, 関節速度が振動する.
  そのため, ノイズを入れない環境ではOurs-1はOurs-2よりも良い性能を示したのに対して, ノイズを入れた環境ではOurs-1はOurs-2だけでなくBasicよりも低い性能となってしまった.
  一方で関節角度は関節速度に比べてノイズが抑えられるため, ノイズを入れてもOurs-2の性能はほとんど落ちなかった.
  そして, この影響は実機においてより顕著に現れた.
  実機におけるワイヤ張力発生時のワイヤ長の振動は想定していたものよりも大きく, Ours-1の場合は一度たりともジャンプすることができなかった.
  関節速度の振動と指令ワイヤ張力の振動が交互に起こり動作が発散してしまう.
  一方で, Ours-2では関節速度の振動に対して大きく影響を受けず, 指令ワイヤ張力も安定しているため, 結果的に関節速度の遷移もOurs-1のときに比べて滑らかであった.
  最終的には4回の連続跳躍に成功している.

  本研究の問題と今後の展望について述べる.
  本研究はワイヤの伸びを有する拮抗ワイヤ駆動型ロボットの動的動作について, 強化学習の利用とその実機適用性という観点で大きな進展を与えた.
  一方で, 未だ完全な連続跳躍動作には成功していない.
  その理由は以下の3つ(1)--(3)であると考えている.
  (1) まず, 身体の摩擦, 特に関節の動作における摩擦項や出力するワイヤ張力の損失がシミュレーションと実機で合っていない点は重要である.
  これは, \cite{hwangbo2019anymal}におけるActuator Netに相当し, 実機のデータを用いたReal2Simが重要である可能性が高い.
  (2) 次に, ガウシアンノイズだけでなく, 定常誤差を状態に加える必要があると考えている.
  センサの取り付け次第では, 単にガウシアンノイズだけではなく, そのノイズに偏りがある場合が多い.
  それらを考慮したノイズ設計, また, 実機の定常誤差をシミュレーションに反映する方法を考える必要がある.
  (3) 最後に, 観測や行動の遅れへの対処が問題である.
  特にRealsense T265からの本体速度出力や, ワイヤ長からの関節角度推定, モータの慣性による電流入力からワイヤ張力発生までの遅延等を考慮する必要がある.
  これらの問題を全て解決して初めて, 実機においても安定的な跳躍動作が実現できるようになると考えている.

}%

\section{CONCLUSION} \label{sec:conclusion}
\switchlanguage%
{%
  In this study, we have developed a parallel wire-driven monopedal robot that can jump continuously using reinforcement learning.
  While parallel wire-driven robots have high jumping ability by gaining a long acceleration distance, they do not have a joint angle sensor in their minimum configuration, and the joint angle estimation from the wire length may oscillate significantly due to the elongation and loosening of the wire.
  To solve this problem, the time series of joint angles is used as the state input instead of the velocity term, which is susceptible to vibration.
  In addition, the definition of the reward function for a monopedal jumping robot and the method of noise generation for the application to the actual robot are presented.
  In the simulation, the results are much better than those of the previous control, and we can show some aspects of its application to the actual robot.
  We hope that this study will contribute to the realization of dynamic motions of parallel wire-driven robots.
}%
{%
  本研究では, パラレルワイヤ駆動型の一本脚跳躍ロボットの強化学習を用いた連続跳躍を実現した.
  パラレルワイヤ駆動型は長い加速距離を稼ぐことで高い跳躍能力を有する一方, 最小構成では関節角度センサを持たず, ワイヤの伸びと高いワイヤ張力によりワイヤ長からの関節角度推定が大きく振動することがある.
  この問題を解決するため, 振動の影響を受けやすい速度項を用いずに, 関節角度の時系列を状態入力とした.
  また, 一本脚跳躍ロボットに向けた報酬関数の定義, 実機への適用に向けたノイズの与え方を示した.
  シミュレーションにおいて, これまでの制御を大幅に上回る結果が得られ, 実機に置いてもその適用の一端を示すことができた.
  本研究がワイヤの伸びを有するパラレルワイヤ駆動型の動的動作実現に向けた一助となれば幸いである.
}%

{
  \bibliographystyle{IEEEtran}
  \bibliography{main}
}

\end{document}